\documentclass[letterpaper, 10 pt, conference]{ieeeconf}  

\usepackage{cite}
\usepackage{mathtools}
\usepackage{amsmath}
\usepackage{amsfonts}
\usepackage{color}
\usepackage{cuted}
\usepackage{multirow}
\usepackage{booktabs}

\usepackage{amsthm}   
\usepackage{booktabs}
\usepackage{makecell}
\usepackage[version=4]{mhchem}

\usepackage{url}
\usepackage{graphicx}
\newtheorem{theorem}{Theorem}

\IEEEoverridecommandlockouts                              

\title{\LARGE \bf
Machine Learning Detection of Lithium Plating in Lithium-ion Cells: A Gaussian Process Approach
}

\author{Ayush Patnaik$^{1}$, Jackson Fogelquist$^{1}$, Adam B. Zufall$^{1}$, Yiwei Ji$^{2}$, \\ Stephen K. Robinson$^{1}$, Peng Bai$^{2}$ and Xinfan Lin$^{1*}$ 
\thanks{$^{1}$A. Patnaik, J. Fogelquist, A. Zufall, S.K. Robinson, and X. Lin are with the Department of Mechanical and Aerospace Engineering, University of California, Davis, CA 95616, USA, corresponding author e-mail: {\tt\small lxflin@ucdavis.edu}}%
\thanks{$^{2}$Y. Ji and P. Bai are with the Department of Energy, Environmental \& Chemical Engineering and the Institute of Materials Science and Engineering, Washington University in St. Louis, St. Louis, MO 63130, USA.}}%

\begin{document}

\maketitle
\thispagestyle{empty}
\pagestyle{empty}

\begin{abstract} 

Lithium plating during fast charging is a critical degradation mechanism that accelerates capacity fade and can trigger catastrophic safety failures. 
Recent work has shown that plating onset can manifest in incremental-capacity analysis as an additional high-voltage feature above 4.0 V, often appearing as a secondary peak or shoulder distinct from the main intercalation peak complex; however, conventional methods for computing $dQ/dV$ rely on finite differencing with filtering, which amplifies sensor noise and introduces bias in feature location. 
In this paper, we propose a Gaussian Process (GP) framework for lithium plating detection by directly modeling the charge-voltage relationship $Q(V)$ as a stochastic process with calibrated uncertainty. 
Leveraging the property that derivatives of GPs remain GPs, we infer $dQ/dV$ analytically and probabilistically from the posterior, enabling robust detection without ad hoc smoothing. 
The framework provides three key benefits: (i) noise-aware inference with hyperparameters learned from data, (ii) closed-form derivatives with credible intervals for uncertainty quantification, and (iii) scalability to online variants suitable for embedded BMS. 
Experimental validation on Li-ion coin cells across a range of C-rates (0.2C-1C) and temperatures (0-40$^\circ$C) demonstrates that the GP-based method reliably resolves distinct high-voltage secondary peak features under low-temperature, high-rate charging, while correctly reporting no features in non-plating cases. 
The concurrence of GP-identified differential features, reduced charge throughput, capacity fade measured via reference performance tests, and post-mortem microscopy confirmation supports the interpretation of these signatures as plating-related, establishing a practical pathway for real-time lithium plating detection.

\end{abstract}

\section{INTRODUCTION}
\label{sec:introduction}

Fast charging is pivotal not only for accelerating electric-vehicle adoption but also for electrified aerospace applications such as satellites, where mission timelines and energy availability demand rapid and reliable charge acceptance \cite{liu_challenges_2019, waldmann_li_2018}. 
In orbit, repeated short sun–eclipse cycles and limited thermal control expose cells to wide temperature swings, and charging under sub-ambient conditions is a well-established trigger for metallic lithium (Li) plating, a detrimental degradation phenomenon \cite{liu_challenges_2019, waldmann_li_2018, ng_low-temperature_2020, petzl_lithium_2015}. 
Accordingly, Li plating is recognized as one of the most critical challenges of fast charging, making its early detection and prevention essential for both safety and lifetime \cite{waldmann_li_2018, ng_low-temperature_2020}.


In the electrochemical research community, lithium plating has been extensively studied using a wide range of laboratory-based diagnostic techniques \cite{waldmann_li_2018}. 
The most straightforward approach is post-mortem analysis, in which the cell is opened and the metallic lithium deposits are directly observed on the electrode surface \cite{McShane2019}. Other in-situ or operando methods such as neutron diffraction, nuclear magnetic resonance (NMR), and X-ray diffraction (XRD) have also been employed to characterize plating and its impact on electrode structure \cite{zinth_lithium_2014, marker_operando_2020, finegan_spatial_2020}.
While powerful, these techniques are either destructive, highly intrusive, or reliant on expensive and complex instrumentation that limits their applicability outside of specialized research laboratories. This has motivated a strong interest in voltage-based diagnostics, which are non-intrusive and thus more suitable for practical battery management in field applications.
Most voltage-based approaches reported to date rely on analysis of relaxation or discharge voltage plateaus associated with stripping of plated lithium \cite{PETZL201480, yang_look_2018, campbell_how_2019}. 
However, these methods only reveal plating after charging has ended and thus do not enable real-time detection during fast-charging operation.
  

Recent studies have reported new voltage features that emerge during charging \cite{ma_operando_2022, chen_operando_2021},  
based on the $dQ/dV$ curve obtained by differentiating the accumulated charge $Q$ (the integral of current) with respect to voltage $V$. 
Under plating-prone conditions, an additional secondary peak distinct from the main intercalation peak complex emerges at high cell voltage near end-of-charge, which has been directly linked to the onset of lithium plating.
Operando experiments using novel capillary cell setups have confirmed this finding by directly visualizing the growth of metallic Li deposits coincident with the appearance of the voltage feature \cite{ma_operando_2022}. 
This discovery opens up a promising pathway toward real-time detection and control of lithium plating: in principle, one could compute $dQ/dV$ from routine current and voltage measurements and terminate charging when the peak is observed. 
However, practical implementation poses significant challenges. 
The peak in the $dQ/dV$ curve reflects the flattening of the voltage trajectory, since mathematically $dQ/dV$ is the inverse of the time derivative of voltage under constant-current charging. 
As is well known in control theory, numerical differentiation of signals (e.g., via finite differencing) significantly amplifies sensor noise, leading to large variance in the estimated derivative. 
While laboratory demonstrations can mitigate this issue with high-precision instrumentation and extensive post-processing (e.g., filtering), robust and accurate detection of $dQ/dV$ peaks based on near-zero voltage derivatives remains challenging in practice.

In this work, we propose a Gaussian Process (GP) framework for lithium plating detection by modeling the charge–voltage relationship $Q(V)$ as a GP, directly using the charging voltage curve as input. 
A key property of GPs is that their derivatives are still GPs and remain jointly distributed with the original process \cite{rasmussen_gaussian_2005, solak_derivative_2002}. 
As a result, $dQ/dV$ can be inferred analytically and probabilistically from the GP posterior, without resorting to ad hoc numerical differencing. 
Compared to the widely adopted filtering–differentiation methods for $dQ/dV$ analysis, such as the Savitzky–Golay (SG) filter \cite{savitzky_smoothing_1964}, the GP approach offers several advantages:  
(i) GP inference provides a principled treatment of measurement noise, where the hyperparameters 
are self-learned from data, while the traditional filtering approach requires manual tuning of window length and filtering parameters;  
(ii) GP derivatives are obtained in closed form, thereby avoiding the variance amplification inherent to finite differencing; and  
(iii) GP yields not only point estimates of $dQ/dV$ but also credible intervals, enabling explicit quantification of uncertainty.  
The probabilistic nature of GPs further makes the method naturally extensible to online detection and adaptive control, where uncertainty-aware decision making is critical. 
We demonstrate the method under fast-charge and low-temperature conditions across no-plating and plating datasets, showing the sensitivity and reliability of GP-based solution in the presence of sensor noise.

\section{Mechanism of Lithium Plating and Electrochemical Features}

Lithium plating occurs when polarization during charging drives the anode potential below 0\,V vs.\ Li/Li$^+$, favoring metallic deposition over intercalation \cite{waldmann_li_2018}.
Incremental capacity analysis ($dQ/dV$) is a standard diagnostic tool whose peaks correspond to electrode phase transitions and shift systematically with degradation \cite{beatty2024review, chen_operando_2021, Dubarry2022_ICA_review}.
Lithium plating, however, differs fundamentally from these slow mechanisms. 
It progresses rapidly and can cause severe capacity loss within only a few charge-discharge cycles. 
Failure to detect plating early not only shortens cell lifetime but also introduces serious safety risks. 
Recent operando studies have shown that under fast charging, the ordinary intercalation peaks in the incremental-capacity spectrum can shift and broaden toward higher cell voltages as polarization increases \cite{chen_operando_2021}. 
Therefore, lithium plating should not be inferred from the absolute location of the dominant peak alone. 
Rather, plating is associated with the emergence of an additional high-voltage feature near end-of-charge, typically a secondary peak, shoulder, or hump superimposed on the main intercalation peak complex \cite{ma_operando_2022, chen_operando_2021}. 
This feature is linked to the transition from lithium intercalation to nucleation and growth on the graphite surface, and correlates with the local minimum in graphite potential during charging \cite{chen_operando_2021}. 
In this work, we therefore use the secondary high-voltage peak as the distinct plating-related feature. 

Despite its promise, applying $dQ/dV$ analysis for plating detection under practical fast-charging conditions remains challenging. 
At high C-rates, overpotentials and ohmic (IR) drop distort the voltage curve, shifting and broadening differential features. 
More critically, numerical differentiation of noisy voltage-current measurements amplifies sensor noise, producing fluctuations that obscure the plating peak. 
Traditional filtering methods such as Savitzky-Golay \cite{savitzky_smoothing_1964, beatty2024review}, can mitigate variance but introduce their own biases, making robust peak detection an open challenge.

\section{Gaussian Process-based Lithium Plating Detection Framework} 


Gaussian Processes (GPs) offer a principled alternative: they model the underlying charge–voltage map as a smooth function with calibrated uncertainty, and enable noise-robust derivative inference via closed-form conditioning rather than unstable differencing \cite{rasmussen_gaussian_2005, solak_derivative_2002, pmlr-v5-titsias09a}. 
Specifically, GPs (i) capture nonlinear yet smooth electrochemical responses; (ii) provide credible intervals that quantify the confidence of plating-related peaks; and (iii) admit scalable variants (e.g., state-space, sparse, or variational GPs) suitable for online deployment in battery management systems (BMS) for both EVs and spacecraft \cite{fogelquist2025combining, richardson_gaussian_2017}.

\subsection{Model Formulation: GP over Q(V)}
We consider a GP model formalism as
\begin{equation}
\begin{split}
    y &= f(x) + \varepsilon, \; \varepsilon \sim \mathcal{N}(0, \sigma_n^2) \\
    f&(x) \sim  \mathcal{GP}(m(x), k(x, x')). 
\end{split}
\label{GP}
\end{equation}
Specifically in our case, 
the input $x$ is the battery voltage $V$, 
the output $y$ is the accumulated charge $Q$ during charging
\begin{equation}
  Q = \int_{0}^t I dt, 
\end{equation}
which is modeled as a function of $V$ in $f(x)$ with additive 
zero-mean Gaussian noise $\varepsilon$ with variance $\sigma_n^2$. 

The function $f(x)$ is characterized by a Gaussian process,
with mean and variance dependent on $x$.
Specifically, it means that any pair of $f(x)$ and $f(x')$ are joint Gaussian 
with respective mean and covariance. 
Here, the mean is set to zero as a common practice,
while the (co-)variance is characterized by a kernel function $k(x,x')$. 
The kernel function encodes the similarity between the input $x$ of interest 
and other (e.g. training) inputs in the space $x'$, 
thereby determining how information from $f(x')$ contributes to the prediction of $f(x)$.
In this work, we use the classical squared-exponential kernel, 
\begin{equation}
k(x, x') = \sigma_f^2 \exp\!\left(-\frac{(x - x')^2}{2\ell^2}\right),
\label{RBF}
\end{equation}
where the length scale $l$ 
controls the smoothness of the function, 
and $\sigma^2_f$ controls the variation among data \cite{rasmussen_gaussian_2005}. 


The hyperparameters of the model, namely $\theta = \{\ell, \sigma_f, \sigma_n\}$,
will be trained by maximizing the log marginal likelihood of the prediction over the training data set,
which takes the following form under our GP formulation \cite{rasmussen_gaussian_2005},
\begin{equation*}
   \max_\theta 
   -\frac{1}{2}Y^\top K_n^{-1}(\theta)Y - \frac{1}{2}\log|K_n(\theta)| - \frac{N}{2}\log(2\pi). 
\end{equation*}
In the equation, the vector ${Y} = [y_1, y_2, \cdots, y_N]$ represents the training output dataset,
${X} = [x_1, x_2, \cdots, x_N]$ consists of the training input data,  
and $K_n = K({X, X}) + \sigma_n^2$ is the covariance matrix over the training data
with each element $[K(X, X]_{ij}=k(x_i, x_j)$ specified by the kernel in Eqn. (\ref{RBF}).
Here $y=f(x)+\varepsilon$, $\varepsilon\sim\mathcal{N}(0,\sigma_n^2)$ induces $K_n(\theta)=K(X,X)+\sigma_n^2 I$, so $\sigma_n$ is learned jointly with $\ell$ and $\sigma_f$ by maximizing the log marginal likelihood \cite{rasmussen_gaussian_2005}.



\subsection{Inference of $dQ/dV$}

Our primary objective is to obtain $dQ/dV$, i.e., the derivative of the output with respect to the input, 
\[
f'(x) = \frac{dy}{dx}.
\]
To this end, we leverage a key property of Gaussian Processes as specified in Theorem 1: the derivative of a GP is itself a GP, and it is jointly Gaussian with the original process \cite{rasmussen_gaussian_2005, solak_derivative_2002}. 
This property allows us to infer $dQ/dV$ directly from data without unstable numerical differencing.
\begin{theorem}[Derivative GP and Joint Gaussianity \cite{rasmussen_gaussian_2005}]
Let 
\[
f \sim \mathcal{GP}\!\big(m(\cdot), k(\cdot,\cdot)\big), 
\quad \mathcal{X} \subset \mathbb{R},
\]
and assume $f$ is mean-square differentiable (equivalently, $k$ is $C^2$ near the diagonal). 
Then the derivative process
\[
f'(x) = \frac{d}{dx} f(x)
\]
is also a Gaussian process with mean and covariance
\[
m'(x) = \frac{d}{dx} m(x), 
\qquad
k'(x,x') = \frac{\partial^2}{\partial x\,\partial x'} k(x,x').
\]
Moreover, $f$ and $f'$ are jointly Gaussian. For any finite sets
\[
X = \{x_1,\dots,x_n\}, 
\quad 
X' = \{x'_1,\dots,x'_m\},
\]
the random vector
\[
\begin{bmatrix}
f(X)\\[2pt] f'(X')
\end{bmatrix}
=
\begin{bmatrix}
f(x_1),\dots,f(x_n),\, f'(x'_1),\dots,f'(x'_m)
\end{bmatrix}^\top
\]
is multivariate normal with mean
\[
\begin{bmatrix}
m(X)\\[2pt] m'(X')
\end{bmatrix}
\]
and block covariance
\[
\begin{bmatrix}
K(X,X) & K'(X,X')\\[4pt]
K'(X',X) & K''(X',X')
\end{bmatrix},
\]
where the blocks are defined element-wise by
\begin{equation}
\begin{split}
\big[K(X,X)\big]_{ij} &= k(x_i,x_j), \\
\big[K'(X,X')\big]_{ij} &= \frac{\partial}{\partial x'_j} k(x_i, x'_j), \\
\big[K'(X',X)\big]_{ij} &= \frac{\partial}{\partial x'_i} k(x'_i, x_j), \\
\big[K''(X',X')\big]_{ij} &= \frac{\partial^2}{\partial x'_i\,\partial x'_j} k(x'_i, x'_j).
\end{split}
\end{equation}
\end{theorem}

The above theorem follows from the fundamental property that any linear operation applied to a Gaussian Process yields another Gaussian Process, and differentiation is a linear operator. 
Therefore, in our setting we can directly formulate a new GP corresponding to the derivative process, tailored to the estimation of $dQ/dV$, 
\begin{equation}
    \begin{bmatrix} Y \\ F'_* \end{bmatrix} \sim \mathcal{N}\left(
    \begin{bmatrix} 0 \\ 0 \end{bmatrix},
    \begin{bmatrix} K(X, X) + \sigma_n^2 \bf{I} & K'({X,X^*})  \\ {K'}({X^*,X}) & K''({X^*,X^*}) \end{bmatrix}
    \right),
\end{equation}
where ${Y} = [f(y_1), f(y_2), \cdots, f(y_N)]$ are the $Q(V)$ data, 
and ${F'(X^*)} = [f'(x^*_1), f'(x^*_2), \cdots, f'(x^*_{M})]$ are the $dQ/dV$ that need to be predicted at the target voltage $V^*_i$'s.
Meanwhile, based on the squared-exponential kernel, the covariance matrices are given as
\begin{equation}
\begin{split}
[K'({X,X^*})]_{ij} &= \partial_{x^*} k(x_i, x^*_j) =  k(x_i, x^*_j) \frac{(x_i-x^*_j)}{\ell^2}\\ 
[K''(X^*,X^*)]_{ij} &=  \partial_x \partial_{x^*} k(x_i, x^*_j) \\ 
&= k(x_i, x^*_j) \left( \frac{1}{\ell^2} - \frac{(x_i-x^*_j)^2}{\ell^4} \right). 
\end{split}
\end{equation}

Finally, based on the new joint Gaussian process, 
we can provide a posterior estimate of the $dQ/dV$ curve ${F'(X^*)}$ conditional on the 
  noisy $\mathbf{y}$ data as 
\begin{equation*}
    F'(x^*) | Y \sim \mathcal{N}(\mu'^*, \Sigma'^*)
\end{equation*}
with
\begin{align*}
    \mu'(X^*) &= K'({X^*,X}) K(X,X)^{-1}Y, \\
    \Sigma'^* &= K''({X^*,X^*}) - K'({X_*,X}) K(X,X)^{-1} K'({X,X_*}).
\end{align*}
where, the mean $\mu'(X^*)$ gives the estimate of the $dQ/dV$,  
while $\Sigma'^*$ can be used to construct the confidence interval of the estimate.
We use kernels with $C^2$ regularity (squared-exponential/RBF), which ensure mean-square differentiability so that the derivative process $f'(x)$ exists and remains GP-distributed; identical results hold for Matérn-$5/2$ \cite{rasmussen_gaussian_2005}.

The GP operates on $Q$ as a function of $V$ and does not require constant current. For multi-stage profiles, we infer $dQ/dV$ directly from the value–derivative posterior without invoking the CC identity $dQ/dV=I/(dV/dt)$; therefore the method applies to monotonic charge segments of arbitrary current profiles.

\subsection{Training/Validation Protocol and Terminology}
We train the GP once, using only the first charge cycle of the $1$C, $10~^{\circ}\mathrm{C}$ case. 
The hyperparameters $\theta=\{\ell,\sigma_f,\sigma_n\}$ are learned by maximum likelihood from that cycle’s $Q(V)$ data and then held fixed for all subsequent analyses; for every other cycle/condition we only condition the GP on that cycle’s $(V,Q)$ pairs to form the posterior (no re-tuning across temperatures or C-rates).

\section{Experimental Validation}

\subsection{Testing Setup}
\begin{figure*}[t]
    \centering
    \includegraphics[width=0.9\textwidth]{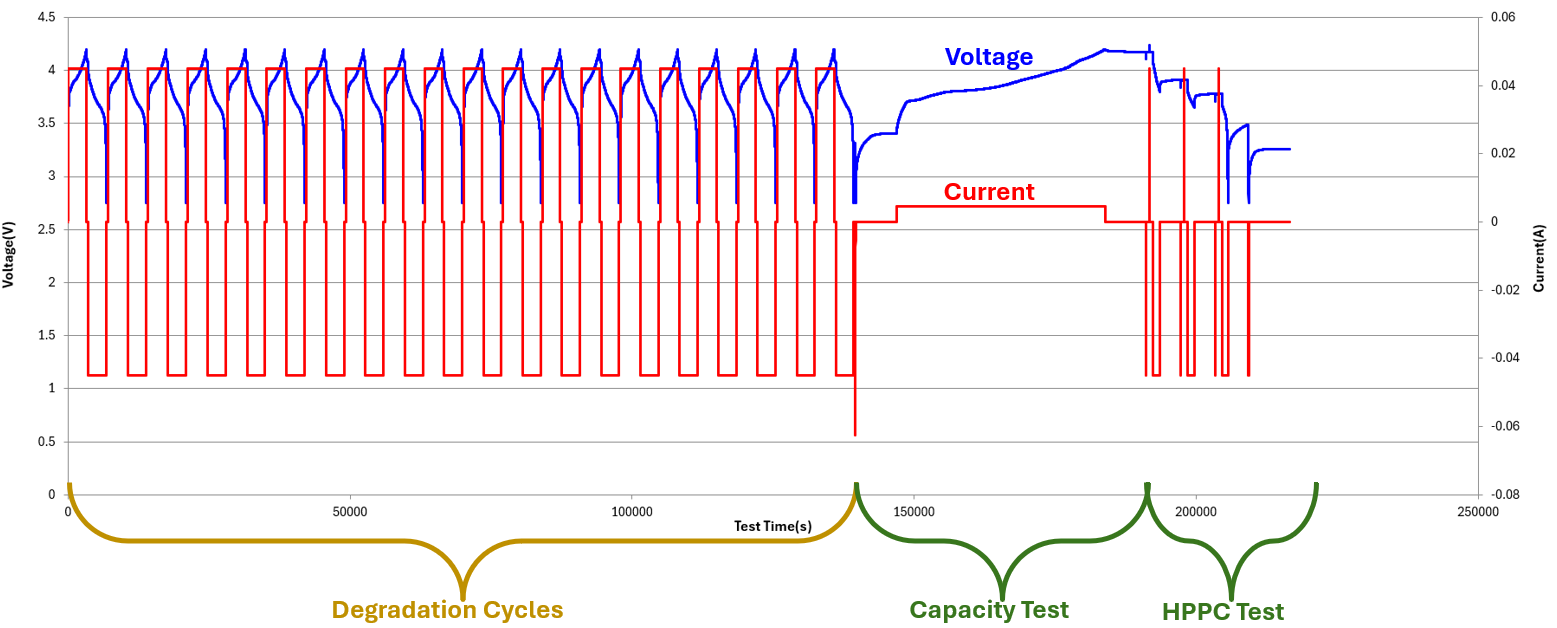}
    \caption{Representative test sequence (1C at $25\,^{\circ}\mathrm{C}$) showing the complete protocol used for experiments. Degradation cycles consisting of repeated constant-current charge/discharge cycles (left), followed by a Reference Performance Test (RPT) consisting of a low-rate capacity test and Hybrid Pulse Power Characterization (HPPC) pulses (right). Blue = voltage (left axis); red = current (right axis).}
    \label{fig:TestingProtocol}
\end{figure*}


To validate the proposed lithium plating detection scheme, we conducted constant-current (CC) charge/discharge aging cycles under multiple operating conditions. 
Charging was performed at C-rates of 1C, 0.6C, 0.4C, and 0.2C, across ambient temperatures ranging from $40\,^{\circ}\text{C}$ down to $0\,^{\circ}\text{C}$ (inclusive). 
Each charging cycle was followed by discharging at 1C down to the manufacturer-specified lower cutoff voltage $V_{\min}=2.75~\text{V}$, with an upper cutoff voltage $V_{\max}=4.2~\text{V}$. 
A rest period of $5~\text{min}$ was inserted at the end of each cycle, and at least 10 cycles were repeated under each condition. 
The experiments were performed using an Arbin LBT 21084 battery cycler for charge/discharge control and data acquisition, together with an ESPEC environmental chamber for temperature regulation. 
The tested cells were EEMB LIR-2032 graphite$\parallel$LiCoO$_2$ coin cells (nominal capacity of $45$ mAh). 
To track degradation, periodic Reference Performance Tests (RPTs) were conducted, including C/10 constant-current capacity tests and hybrid pulse power characterization (HPPC) to evaluate internal resistance.
Cells were also disassembled after cycling for post-mortem optical microscopy (Olympus BX53M) of the graphite anodes to provide independent physical verification of plating classifications.
A typical testing sequence is shown in Fig. \ref{fig:TestingProtocol}.

\subsection{Results and Discussion}


\begin{figure*}[t]
    \centering
    \includegraphics[width=0.9\textwidth]{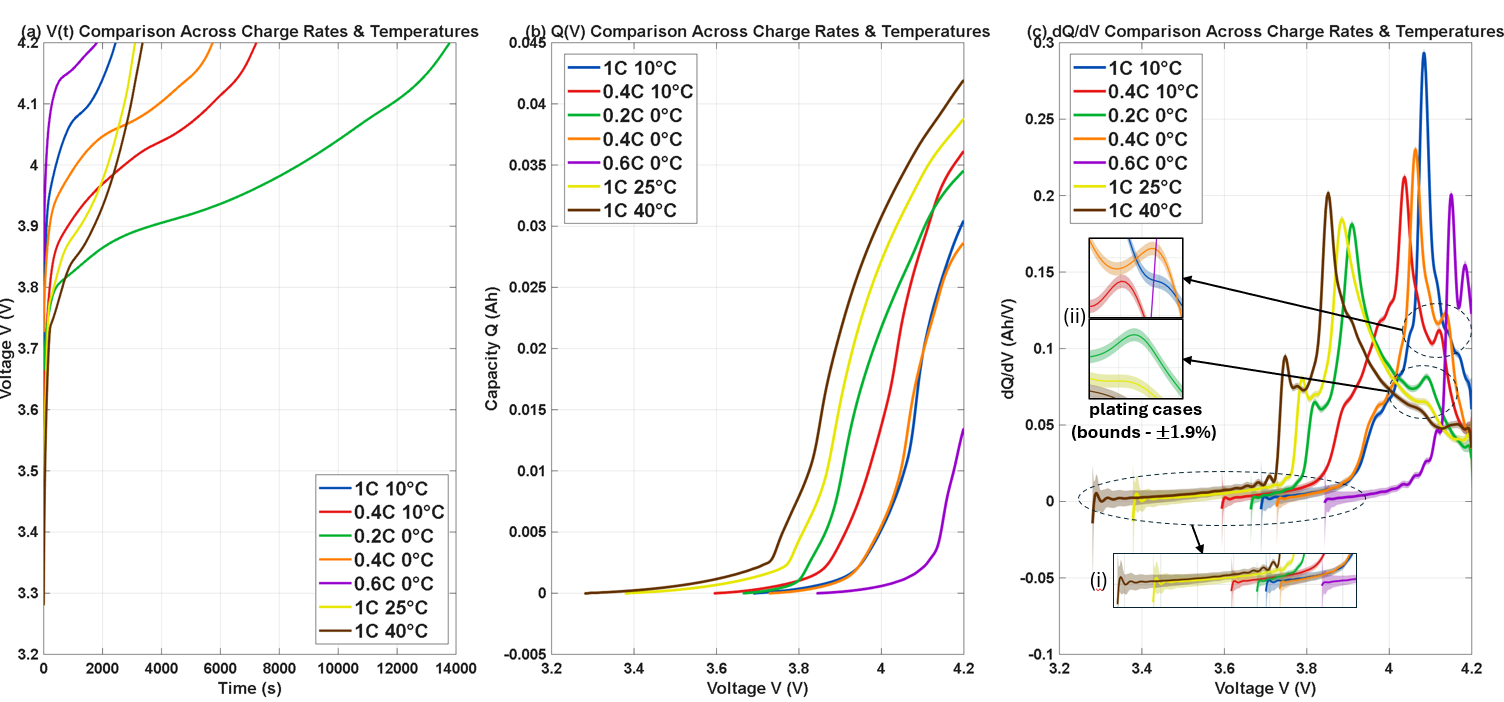}
    \caption{GP-based plating detection across charge rates and temperatures. (a) V(t) during CC charge for representative cycles (legend shows C-rate and temperature). (b) Capacity–voltage trajectories Q(V). (c) GP-derived incremental capacity $dQ/dV$ with a secondary peak above 4 V distinguishing plating from no-plating conditions. Inset figures (i) and (ii) show magnified 95\% credible bands at the start of the cycle and at the secondary peak respectively, with bounds expressed as ±1.9\% of the peak magnitude of all cases and the plating cluster respectively.}
    \label{fig:GPResults}
\end{figure*}



\begin{figure*}[t]
    \centering
    \includegraphics[width=0.33\textwidth]{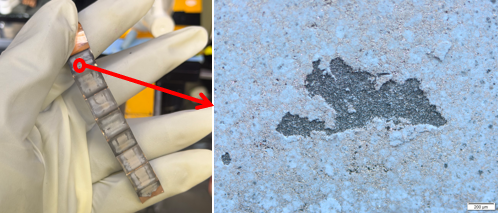}%
    \includegraphics[width=0.33\textwidth]{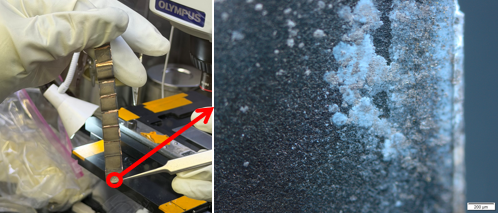}%
    \includegraphics[width=0.33\textwidth]{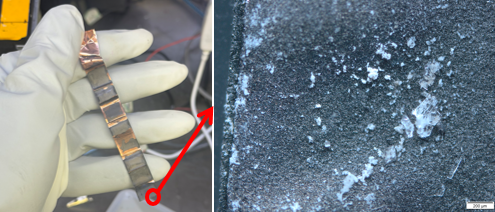}
    
    \makebox[0.33\textwidth]{(a)}%
    \makebox[0.33\textwidth]{(b)}%
    \makebox[0.33\textwidth]{(c)}
    
    \caption{Post-mortem optical microscopy of graphite anodes after cycling (scale bar: $200\,\mu\text{m}$). 
    (a) 0.4C at $10\,^{\circ}\text{C}$: uniform metallic lithium deposits across the anode surface, indicating severe plating. 
    (b) 1C at $25\,^{\circ}\text{C}$: sparse, localized metallic lithium deposits, indicating mild or onset-stage plating. 
    (c) 1C at $10\,^{\circ}\text{C}$: no metallic lithium observed; white crystalline residues with organic morphology suggesting electrolyte decomposition products.} 
    \label{fig:PostMortem}
\end{figure*}

Using voltage–current data collected during charging, we trained GP models on the measured $Q(V)$ trajectory and inferred $dQ/dV$ directly from the derivative GP posterior. 
This probabilistic differentiator yields smooth incremental-capacity curves with calibrated uncertainty, enabling robust peak finding under realistic sensor noises. 
Fig. \ref{fig:GPResults} illustrates the full GP-based analysis workflow. 
The first subplot shows $V(t)$ during CC charge, where increasing C-rate and decreasing temperature shorten charge time due to increased polarization. 
The second subplot maps the data to $Q(V)$ and overlays the GP posterior.
The third subplot presents the GP-derived $dQ/dV$ (mean and $\pm$95\% credible interval), revealing transient differential peaks otherwise obscured by finite-differencing.
Post-mortem optical microscopy results are shown in Fig. \ref{fig:PostMortem}, which provides the direct evidence for validating lithium plating. 
For example, the anode tested under 0.4C $10\,^{\circ}\text{C}$ (Fig. \ref{fig:PostMortem}a) exhibits uniform metallic lithium deposits covering the electrode surface, indicating severe plating. 
The anode tested under $25\,^{\circ}\text{C}$ anode (Fig. \ref{fig:PostMortem}b) shows sparse, localized metallic deposits, suggesting mild or onset-stage plating. 
The full microscopy results gave the lithium plating labels in Tab. \ref{tab:results-summary}, which also summarizes all experimental validation results. 
        

\begin{table}[htbp]
    \centering
    \caption{Experimental Validation Results}
    \label{tab:results-summary}
    \setlength{\tabcolsep}{3pt} 
    \begin{tabular}{lcccc} 
        \toprule
        \thead{Operating \\ condition} & \thead{Second Peak \\ $>$ 4.0 V?} & \thead{Lithium \\ plating?} & \thead{Charge throughput \\ decrease rate \\ (\% loss/cycle)} & \thead{RPT Capacity \\ decrease rate \\ (\% loss/cycle)}
        \\
        
        \midrule
        $1.0\,\mathrm{C}$@$10^{\circ}\mathrm{C}$& No & No & 1.464 & 0.877 \\
        $0.4\,\mathrm{C}$@$10^{\circ}\mathrm{C}$& Yes & Yes & 1.671 & 1.237 \\
        $0.6\,\mathrm{C}$@$0^{\circ}\mathrm{C}$& Yes & Yes & 3.617 & 1.175 \\
        $0.4\,\mathrm{C}$@$0^{\circ}\mathrm{C}$& Yes & Yes & 1.711 & 0.845 \\
        $0.2\,\mathrm{C}$@$0^{\circ}\mathrm{C}$&  Yes  & Yes  & 0.029 & 0.094 \\
        $1.0\,\mathrm{C}$@$40^{\circ}\mathrm{C}$   & No  & No  & 0.093 & 0.016 \\
        $1.0\,\mathrm{C}$@$25^{\circ}\mathrm{C}$   & Yes  & Yes  & 0.227 & 0.031 \\
        \bottomrule
    \end{tabular}
\end{table}
Fig. \ref{fig:GPResults} shows that, across the tested operating conditions, the GP-based $dQ/dV$ analysis reveals a clear difference between the plating and no-plating cases: 
one cluster exhibits a distinct secondary peak above $4.0~\mathrm{V}$, while the other shows no such peak. 
According to prior studies \cite{ma_operando_2022}, a secondary peak above $4.0~\mathrm{V}$ is the characteristic electrochemical signature of lithium plating. 
Closer inspection reveals that the presence or absence of a secondary peak does not follow a simple rate–temperature rule. For instance, 1C at $25\,^{\circ}\text{C}$
exhibits a secondary peak despite being at room temperature, while 1C at $10\,^{\circ}\text{C}$ does not, even though the latter operates at a lower temperature. 
Similarly, 0.2C at $0\,^{\circ}\text{C}$ shows a clear secondary peak at a very low charging rate.
These observations suggest that the interplay between current density, temperature, and electrode-level transport determines whether the anode potential locally crosses the plating threshold and that the GP-resolved feature captures this transition regardless of whether the operating condition would be classified as 'aggressive' by conventional heuristics. 
In contrast, the two cases without a secondary peak: 1C at $10\,^{\circ}\text{C}$ and 1C at $40\,^{\circ}\text{C}$ span a wide temperature range, indicating that the absence of a secondary feature reflects the specific electrochemical state at the anode rather than charging rate and temperature alone.


Notably, the 1C at $10\,^{\circ}\text{C}$ condition exhibits no secondary peak yet shows the highest capacity decrease rate among all cases (0.877\%/cycle from RPT). 
Post-mortem optical microscopy of this cell (Fig. \ref{fig:PostMortem}c) reveals no metallic lithium deposits on the graphite surface; instead, white crystalline deposits of organic morphology are observed, consistent with electrolyte decomposition products precipitated under elevated anode polarization at low temperature. 
This is in agreement with the understanding that capacity fade at low temperature can also be caused by non-plating parasitic mechanisms, e.g. electrolyte reduction and SEI growth \cite{inl_cycle_evolution, petzl_lithium_2015, waldmann_li_2018}. 
The absence of both a secondary $dQ/dV$ peak and physical lithium deposits in this case, despite significant capacity fade, confirms that the GP-identified plating signature exhibits specificity: it distinguishes lithium plating from other degradation pathways that also consume lithium.

To further examine the consistency of these classifications, we also tracked two other degradation metrics: charge throughput of charging cycles and capacity measured via RPTs. 
Most conditions classified as plating-positive by the secondary-peak criterion also exhibited elevated throughput loss and capacity fade, with rates roughly an order of magnitude larger than those of the feature-negative cases (Table~\ref{tab:results-summary}). 
However, the correspondence is not one-to-one: the 0.2C at $0\,^{\circ}\text{C}$ condition exhibits a clear secondary peak yet shows only minimal throughput decrease (0.029\%/cycle) and capacity fade (0.094\%/cycle), comparable to the non-plating cases. 
This is consistent with the interpretation that the secondary peak captures the onset of plating, a regime in which lithium loss has not yet accumulated to levels detectable by bulk performance metrics. 
Similarly, 1C at $25\,^{\circ}\text{C}$ shows a secondary peak with modest degradation rates (0.227\%/cycle throughput, 0.031\%/cycle capacity), suggesting early or mild plating that conventional methods would likely miss.


Finally, it is important to note the role of GP uncertainty quantification. 
Across all plating cases, the credible intervals around the detected peaks were exceptionally tight
as shown in the insets of Fig. \ref{fig:GPResults}(c). 
On average, the bounds were within 1.9\% of the peak magnitude, indicating very high confidence in the detection. 
This level of certainty is unattainable with conventional methods, which provide only smoothed (point) estimates without any measure of confidence.

\section{Conclusions}

We presented a voltage-base method for lithium plating detection that employs Gaussian Process. 
By analytically computing $dQ/dV$ from voltage and current using the joint value–derivative posterior of GP, the method avoids the noise amplification inherent to finite-difference ICA/DVA and provides calibrated uncertainty as a reliability metric. 
Across the tested conditions, the GP-based analysis separates cases that exhibit a distinct high-voltage secondary peak feature from cases that do not. 
Feature-positive cases span a range of conditions, including room-temperature charging at 1C while the feature-negative cases 
demonstrate that the absence of plating depends on the specific electrochemical state rather than a simple charging rate–temperature threshold.


The GP-resolved secondary $dQ/dV$ peak above 4.0 V serves as the primary indicator of plating onset, with reduced charge throughput and accelerated capacity fade providing corroborating evidence in cases where plating has progressed. 
Notably, the method detects plating even in conditions such as 0.2C at $0\,^{\circ}\text{C}$ or 1C at $25\,^{\circ}\text{C}$ where degradation metrics remain near non-plating levels, demonstrating the sensitivity of the GP-based approach as an early-warning diagnostic.
Moreover, the method demonstrates specificity as well as sensitivity: the 1C at $10\,^{\circ}\text{C}$ condition, which exhibited significant capacity fade and whose post-mortem analysis revealed electrolyte decomposition products but no metallic lithium, correctly returned no plating feature, confirming that the GP-identified signature discriminates lithium plating from other degradation pathways.
These results demonstrate that the GP–based voltage analysis enables more robust lithium plating diagnostics, paving the way for reliable online detection and integration into advanced battery management systems.

\addtolength{\textheight}{-4cm}   





\section*{ACKNOWLEDGMENT}

We appreciate the funding support from the AFRL Proteus Space STTR and the NSF CAREER Program (No. 2044932). 


\bibliographystyle{IEEEtran}
\bibliography{Reference}

@article{liu_challenges_2019,
	title = {Challenges and opportunities towards fast-charging battery materials},
	volume = {4},
	rights = {2019 Springer Nature Limited},
	issn = {2058-7546},
	url = {},
	doi = {10.1038/s41560-019-0405-3},
	pages = {540--550},
    year = {2019},
	number = {7},
	journal = {Nature Energy},
	shortjournal = {Nat Energy},
	author = {Liu, Yayuan and Zhu, Yangying and Cui, Yi},
	urldate = {2025-09-11},
	date = {2019-07},
	langid = {english},
	note = {},
}

@inproceedings{solak_derivative_2002,
 author = {Solak, E. and Murray-smith, R. and Leithead, W. and Leith, D. and Rasmussen, Carl},
 booktitle = {Advances in Neural Information Processing Systems},
 editor = {S. Becker and S. Thrun and K. Obermayer},
 pages = {},
 publisher = {MIT Press},
 title = {Derivative Observations in Gaussian Process Models of Dynamic Systems},
 url = {},
 volume = {14},
 year = {2001}
}

@InProceedings{pmlr-v5-titsias09a,
  title = 	 {Variational Learning of Inducing Variables in Sparse Gaussian Processes},
  author = 	 {Titsias, Michalis},
  booktitle = 	 {Proceedings of the Twelfth International Conference on Artificial Intelligence and Statistics},
  pages = 	 {567--574},
  year = 	 {2009},
  editor = 	 {van Dyk, David and Welling, Max},
  volume = 	 {5},
  series = 	 {Proceedings of Machine Learning Research},
  address = 	 {},
  month = 	 {16--18 Apr},
  publisher =    {PMLR},
  url = 	 {}
}

@article{waldmann_li_2018,
	title = {Li plating as unwanted side reaction in commercial Li-ion cells – A review},
	volume = {384},
	issn = {0378-7753},
	url = {},
	doi = {10.1016/j.jpowsour.2018.02.063},
	pages = {107--124},
    year = {2018},
	journal = {Journal of Power Sources},
	shortjournal = {Journal of Power Sources},
	author = {Waldmann, Thomas and Hogg, Björn-Ingo and Wohlfahrt-Mehrens, Margret},
	urldate = {2025-09-11},
	date = {2018-04-30},
}

@article{ng_low-temperature_2020,
	title = {Low-Temperature Lithium Plating/Corrosion Hazard in Lithium-Ion Batteries: Electrode Rippling, Variable States of Charge, and Thermal and Nonthermal Runaway},
	volume = {3},
	url = {},
	doi = {10.1021/acsaem.0c00130},
	shorttitle = {Low-Temperature Lithium Plating/Corrosion Hazard in Lithium-Ion Batteries},
	pages = {3653--3664},
    year = {2018},
	number = {4},
	journal = {{ACS} Applied Energy Materials},
	shortjournal = {{ACS} Appl. Energy Mater.},
	author = {Ng, Benjamin and Coman, Paul T. and Faegh, Ehsan and Peng, Xiong and Karakalos, Stavros G. and Jin, Xinfang and Mustain, William E. and White, Ralph E.},
	urldate = {2025-09-11},
	date = {2020-04-27},
	note = {}}

@article{chen_operando_2021,
	title = {Operando video microscopy of Li plating and re-intercalation on graphite anodes during fast charging},
	volume = {9},
	issn = {2050-7496},
	url = {},
	doi = {10.1039/D1TA06023F},
	pages = {23522--23536},
    year = {2021},
	number = {41},
	journal = {Journal of Materials Chemistry A},
	shortjournal = {J. Mater. Chem. A},
	author = {Chen, Yuxin and Chen, Kuan-Hung and Sanchez, Adrian J. and Kazyak, Eric and Goel, Vishwas and Gorlin, Yelena and Christensen, Jake and Thornton, Katsuyo and Dasgupta, Neil P.},
	urldate = {2025-09-11},
	date = {2021-10-26},
	langid = {english},
	note = {},
}

@article{ma_operando_2022,
	title = {Operando Microscopy Diagnosis of the Onset of Lithium Plating in Transparent Lithium-Ion Full Cells},
	volume = {14},
	issn = {1944-8244},
	url = {},
	doi = {10.1021/acsami.2c16090},
	pages = {54708--54715},
    year = {2022},
	number = {49},
	journal = {{ACS} Applied Materials \& Interfaces},
	shortjournal = {{ACS} Appl. Mater. Interfaces},
	author = {Ma, Bingyuan and Agrawal, Shubham and Gopal, Rajeev and Bai, Peng},
	urldate = {2025-09-11},
	date = {2022-12-14},
	note = {},
}

@article{zinth_lithium_2014,
	title = {Lithium plating in lithium-ion batteries at sub-ambient temperatures investigated by in situ neutron diffraction},
	volume = {271},
	issn = {0378-7753},
	url = {},
	doi = {10.1016/j.jpowsour.2014.07.168},
	pages = {152--159},
    year = {2014},
	journal = {Journal of Power Sources},
	shortjournal = {Journal of Power Sources},
	author = {Zinth, Veronika and von Lüders, Christian and Hofmann, Michael and Hattendorff, Johannes and Buchberger, Irmgard and Erhard, Simon and Rebelo-Kornmeier, Joana and Jossen, Andreas and Gilles, Ralph},
	urldate = {2025-09-11},
	date = {2014-12-20},
}

@article{marker_operando_2020,
	title = {Operando {NMR} of {NMC}811/Graphite Lithium-Ion Batteries: Structure, Dynamics, and Lithium Metal Deposition},
	volume = {142},
	issn = {0002-7863},
	url = {},
	doi = {10.1021/jacs.0c06727},
	shorttitle = {Operando {NMR} of {NMC}811/Graphite Lithium-Ion Batteries},
	pages = {17447--17456},
    year = {2020},
	number = {41},
	journal = {Journal of the American Chemical Society},
	shortjournal = {J. Am. Chem. Soc.},
	author = {Märker, Katharina and Xu, Chao and Grey, Clare P.},
	urldate = {2025-09-11},
	date = {2020-10-14},
	note = {},
}

@article{PETZL201480,
title = {Nondestructive detection, characterization, and quantification of lithium plating in commercial lithium-ion batteries},
journal = {Journal of Power Sources},
volume = {254},
pages = {80-87},
year = {2014},
issn = {0378-7753},
doi = {https://doi.org/10.1016/j.jpowsour.2013.12.060},
url = {},
author = {Mathias Petzl and Michael A. Danzer}
}

@article{fogelquist2025combining,
  title={Combining electrochemistry and data-sparse Gaussian process regression for lithium-ion battery hybrid modeling},
  author={Fogelquist, Jackson and Lin, Xinfan},
  journal={Applied Energy},
  volume={399},
  pages={126458},
  year={2025},
  publisher={Elsevier}
}

@article{beatty2024review,
  title={A review of methods of generating incremental capacity--differential voltage curves for battery health determination},
  author={Beatty, Matthew and Strickland, Dani and Ferreira, Pedro},
  journal={Energies},
  volume={17},
  number={17},
  pages={4309},
  year={2024},
  publisher={MDPI}
}

@article{petzl_lithium_2015,
	title = {Lithium plating in a commercial lithium-ion battery – A low-temperature aging study},
	volume = {275},
	issn = {0378-7753},
	url = {},
	doi = {10.1016/j.jpowsour.2014.11.065},
	pages = {799--807},
    year = {2015},
	journal = {Journal of Power Sources},
	shortjournal = {Journal of Power Sources},
	author = {Petzl, Mathias and Kasper, Michael and Danzer, Michael A.},
	urldate = {2025-09-11},
	date = {2015-02-01},
}

@article{campbell_how_2019,
	title = {How Observable Is Lithium Plating? Differential Voltage Analysis to Identify and Quantify Lithium Plating Following Fast Charging of Cold Lithium-Ion Batteries},
	volume = {166},
	issn = {1945-7111},
	url = {},
	doi = {10.1149/2.0821904jes},
	shorttitle = {How Observable Is Lithium Plating?},
	pages = {A725},
    year = {2019},
	number = {4},
	journal = {Journal of The Electrochemical Society},
	shortjournal = {J. Electrochem. Soc.},
	author = {Campbell, Ian D. and Marzook, Mohamed and Marinescu, Monica and Offer, Gregory J.},
	date = {2019-03-06},
	langid = {english},
	note = {},
}

@article{savitzky_smoothing_1964,
	title = {Smoothing and Differentiation of Data by Simplified Least Squares Procedures.},
	volume = {36},
	issn = {0003-2700},
	url = {},
	doi = {10.1021/ac60214a047},
	pages = {1627--1639},
    year = {1964},
	number = {8},
	journal = {Analytical Chemistry},
	shortjournal = {Anal. Chem.},
	author = {Savitzky, Abraham. and Golay, M. J. E.},
	urldate = {2025-09-11},
	date = {1964-07-01},
	note = {},
	}

@book{rasmussen_gaussian_2005,
	title = {Gaussian Processes for Machine Learning},
	isbn = {978-0-262-25683-4},
	url = {},
	publisher = {The {MIT} Press},
	author = {Rasmussen, Carl Edward and Williams, Christopher K. I.},
	year = {2005-11},
	doi = {10.7551/mitpress/3206.001.0001},
	note = {},
}

@article{richardson_gaussian_2017,
	title = {Gaussian process regression for forecasting battery state of health},
	volume = {357},
	issn = {0378-7753},
	url = {},
	doi = {10.1016/j.jpowsour.2017.05.004},
	pages = {209--219},
    year = {2017},
	journal = {Journal of Power Sources},
	shortjournal = {Journal of Power Sources},
	author = {Richardson, Robert R. and Osborne, Michael A. and Howey, David A.},
	urldate = {2025-09-11},
	date = {2017-07-31},
}

@article{yang_look_2018,
	title = {A look into the voltage plateau signal for detection and quantification of lithium plating in lithium-ion cells},
	volume = {395},
	issn = {0378-7753},
	url = {},
	doi = {10.1016/j.jpowsour.2018.05.073},
	pages = {251--261},
    year = {2018},
	journal = {Journal of Power Sources},
	shortjournal = {Journal of Power Sources},
	author = {Yang, Xiao-Guang and Ge, Shanhai and Liu, Teng and Leng, Yongjun and Wang, Chao-Yang},
	urldate = {2025-09-11},
	date = {2018-08-15},
}

@article{finegan_spatial_2020,
	title = {Spatial dynamics of lithiation and lithium plating during high-rate operation of graphite electrodes},
	volume = {13},
	issn = {1754-5706},
	url = {},
	doi = {10.1039/D0EE01191F},
	pages = {2570--2584},
    year = {2020},
	number = {8},
	journal = {Energy \& Environmental Science},
	shortjournal = {Energy Environ. Sci.},
	author = {Finegan, Donal P. and Quinn, Alexander and Wragg, David S. and Colclasure, Andrew M. and Lu, Xuekun and Tan, Chun and Heenan, Thomas M. M. and Jervis, Rhodri and Brett, Dan J. L. and Das, Supratim and Gao, Tao and Cogswell, Daniel A. and Bazant, Martin Z. and Michiel, Marco Di and Checchia, Stefano and Shearing, Paul R. and Smith, Kandler},
	date = {2020-08-13},
	langid = {english},
	note = {},
	}

@article{McShane2019,
author = {McShane, Eric J. and Colclasure, Andrew M. and Brown, David E. and Konz, Zachary M. and Smith, Kandler and McCloskey, Bryan D.},
title = {Quantification of Inactive Lithium and Solid–Electrolyte Interphase Species on Graphite Electrodes after Fast Charging},
journal = {ACS Energy Letters},
volume = {4},
number = {5},
pages = {2044-2051},
year = {2019},
doi = {9.1021/acsenergylett.0c00859},

URL = {},
eprint = { 
    
        https://doi.org/9.1021/acsenergylett.0c00859
    
    

}

}

@article{Dubarry2022_ICA_review,
    
AUTHOR={Dubarry, Matthieu  and Anseán, David },
           
TITLE={Best practices for incremental capacity analysis},
          
JOURNAL={Frontiers in Energy Research},
          
VOLUME={Volume 10 - 2022},
  
YEAR={2022},
  
URL={},
  
DOI={10.3389/fenrg.2022.1023555},
  
ISSN={2296-598X}}

@article{inl_cycle_evolution,
title = {Sensitivity and reliability of key electrochemical markers for detecting lithium plating during extreme fast charging},
journal = {Journal of Energy Storage},
volume = {46},
pages = {103782},
year = {2022},
issn = {2352-152X},
doi = {https://doi.org/10.1016/j.est.2021.103782},
url = {},
author = {Parameswara R. Chinnam and Tanvir R. Tanim and Eric J. Dufek and Charles C. Dickerson and Meng Li}
}

\end{document}